\documentclass[sigconf]{acmart}
\usepackage{multirow}
\usepackage{tcolorbox} 
\usepackage{multicol}
\usepackage{bm}
\usepackage{algorithm}
\usepackage{algorithmicx}
\usepackage{multirow}
\usepackage{fontawesome}
\usepackage[normalem]{ulem}
\useunder{\uline}{\ul}{}
\allowdisplaybreaks
\usepackage{enumitem}
\setitemize{leftmargin=*}
\usepackage{float}
\usepackage{xcolor}

\AtBeginDocument{%
  \providecommand\BibTeX{{%
    \normalfont B\kern-0.5em{\scshape i\kern-0.25em b}\kern-0.8em\TeX}}}

\setcopyright{acmcopyright}
\copyrightyear{2024}
\acmYear{2024}
\acmDOI{10.1145/1122445.1122456}

\begin{document}

\title{Large Language Model with Graph Convolution for Recommendation}

\author{Yingpeng Du}
\email{dyp1993@pku.edu.cn}
\affiliation{  \institution{Nanyang Technological University}  \country{Singapore}}
\author{Ziyan Wang}
\email{wang1753@e.ntu.edu.sg}
\affiliation{  \institution{Nanyang Technological University}\country{Singapore}}
\author{Zhu Sun}
\email{sunzhuntu@gmail.com	}
\affiliation{\institution{Agency for Science, Technology and Research}\country{Singapore}}
\author{Haoyan Chua}
\email{haoyan001@e.ntu.edu.sg}
\affiliation{\institution{Nanyang Technological University}\country{Singapore}}
\author{Hongzhi Liu}
\email{liuhz@pku.edu.cn}
\affiliation{\institution{Peking University}\country{China}}
\author{Zhonghai Wu}
\email{zhwu@ss.pku.edu.cn}
\affiliation{\institution{Peking University}\country{China}}
\author{Yining Ma}
\email{yiningma@u.nus.edu}
\affiliation{  \institution{Nanyang Technological University}\country{Singapore}}
\author{Jie	Zhang}
\email{zhangj@ntu.edu.sg}
\affiliation{  \institution{Nanyang Technological University}\country{Singapore}}
\author{Youchen Sun}
\email{YOUCHEN001@e.ntu.edu.sg}
\affiliation{  \institution{Nanyang Technological University}\country{Singapore}}

\renewcommand{\shortauthors}{xxx, et al.}

\begin{abstract}

In recent years, efforts have been made to use text information for better user profiling and item characterization in recommendations. However, text information can sometimes be of low quality, hindering its effectiveness for real-world applications. With knowledge and reasoning capabilities capsuled in Large Language Models (LLMs), utilizing LLMs emerges as a promising way for description improvement. However, existing ways of prompting LLMs with raw texts ignore structured knowledge of user-item interactions, which may lead to hallucination problems like inconsistent description generation.  To this end, we propose a {G}raph-{a}ware {C}onvolutional {LLM} method to elicit LLMs to capture high-order relations in the user-item graph. 
 To adapt text-based LLMs with structured graphs,  We use the LLM as an aggregator in graph processing, allowing it to understand graph-based information step by step. Specifically, the LLM is required for description enhancement by exploring multi-hop neighbors layer by layer, thereby propagating information progressively in the graph. To enable LLMs to capture large-scale graph information, we break down the description task into smaller parts, which drastically reduces the context length of the token input with each step. Extensive experiments on three real-world datasets show that our method consistently outperforms state-of-the-art methods. 

\end{abstract}


 \maketitle

 \begin{figure} \centering
 \includegraphics[width=8.5cm]{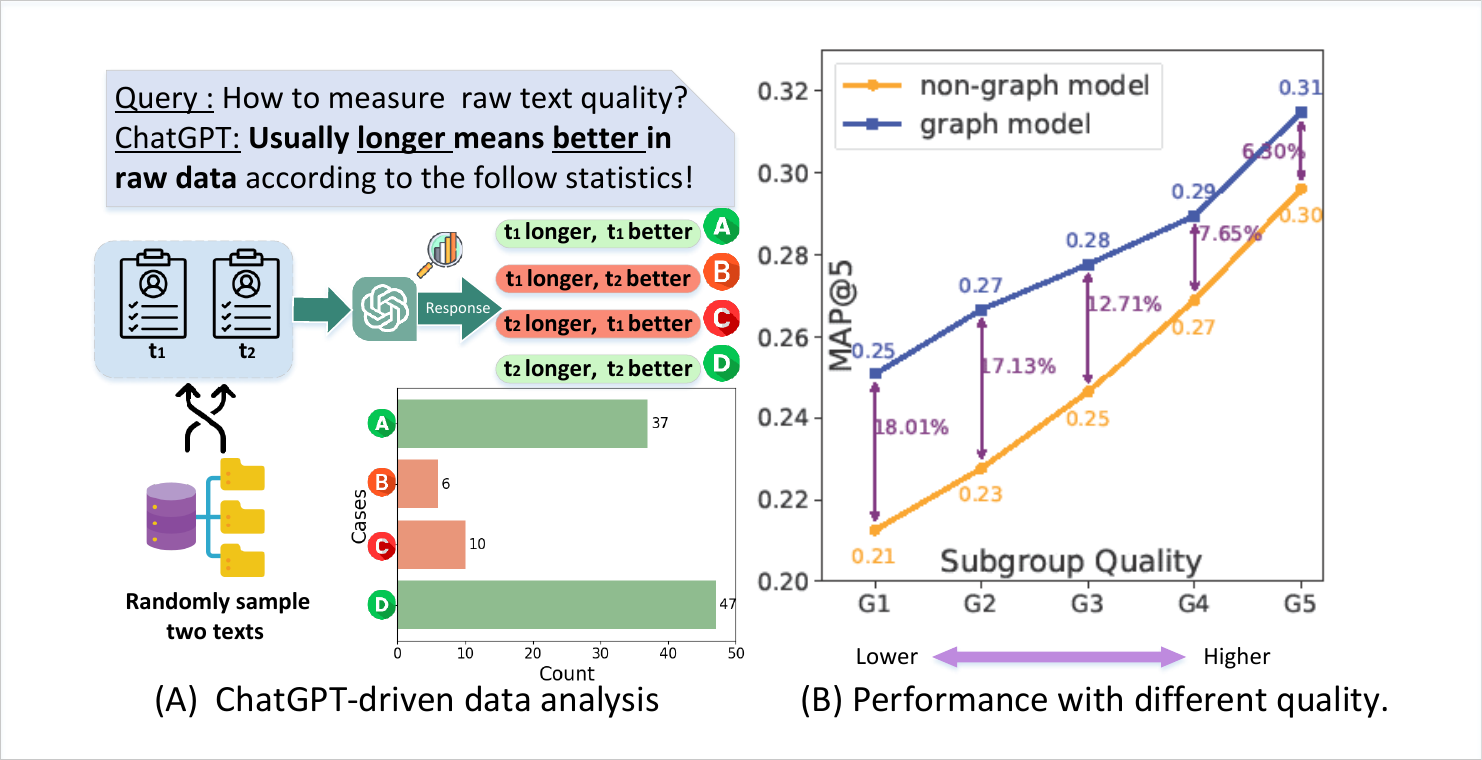}
\caption{ (A) The ChatGPT-driven data analysis. (B) Performance of recommendation methods on user subgroups with different description quality levels.} \label{fig1}
\end{figure}

\section{Introduction} \label{Seuntro}
Recommender systems (RSs) are pivotal in delivering personalized services to users, which can enhance both user satisfaction and platform profitability. Most previous RS methods, e.g. Matrix Factorization~\cite{koren2009matrix}, model users’ preferences and items’ characteristics using merely user-item interaction records, which suffers from data sparsity and leads to suboptimal performance. Nowadays, there has been a trend towards leveraging auxiliary textual information~\cite{kanwal2021review}, such as descriptions of users and items, to achieve better profiling of users and items for recommendation purposes. Nonetheless, due to the unawareness or impatience of users and item providers, the raw descriptions submitted by them may be of low quality. For instance, some users may not carefully craft their self-descriptions or have limited self-awareness of their preferences and traits, leading to incomplete and unreflective descriptions. Such low-quality descriptions hamper the accuracy of recommendation results.

Intuitively, the higher-quality descriptions contribute to better recommendation results. To validate this intuition, we investigate the performance of recommendation methods on user subgroups with different description quality levels, measured by ChatGPT-driven data analysis\footnote{We randomly compare two raw texts submitted by users and ask ChatGPT
to determine which one is better. ChatGPT generally deems the longer one to be more effective (84/100 cases supporting
this idea), which is consistent with our manual evaluation.} as shown in Figure \ref{fig1}(A). It indicates that recommendation methods tend to yield more accurate predictions for users with higher-quality descriptions as shown in Figure \ref{fig1}(B). As a result, considerable attention has been focused on improving and refining raw descriptions of users and items with the help of large language models (LLMs), which have extensive knowledge and inference capabilities. Some approaches rely on directly summarizing raw descriptions or attributes~\cite{zheng2023generative,wu2023towards,wang2023recmind,liu2023chatgpt}, providing only limited improvements due to the
hallucination effect of LLMs~\cite{du2023enhancing,azamfirei2023large}, e.g., inconsistent description generation w.r.t. users' behaviors. To this end, other approaches attempt to incorporate descriptions of engaged items as supplemental knowledge for LLMs~\cite{du2023enhancing,liu2023first}, but the scope of the leveraged information in these methods remains constrained within their adjacent neighbors. Diving deeper into our findings in Figure \ref{fig1} (B), graph-based model~\cite{he2020lightgcn} not only consistently outperforms the non-graph model~\cite{BPR} but also achieves larger improvements on user subgroups with the lower-quality description. These results inspire us to integrate graph information into LLMs to enhance the quality of textual descriptions, i.e., using LLMs to predict and discover the missing descriptions of users and items based on structural graphs, for better recommendation results.

\begin{figure} \centering
 \includegraphics[width=8.5cm]{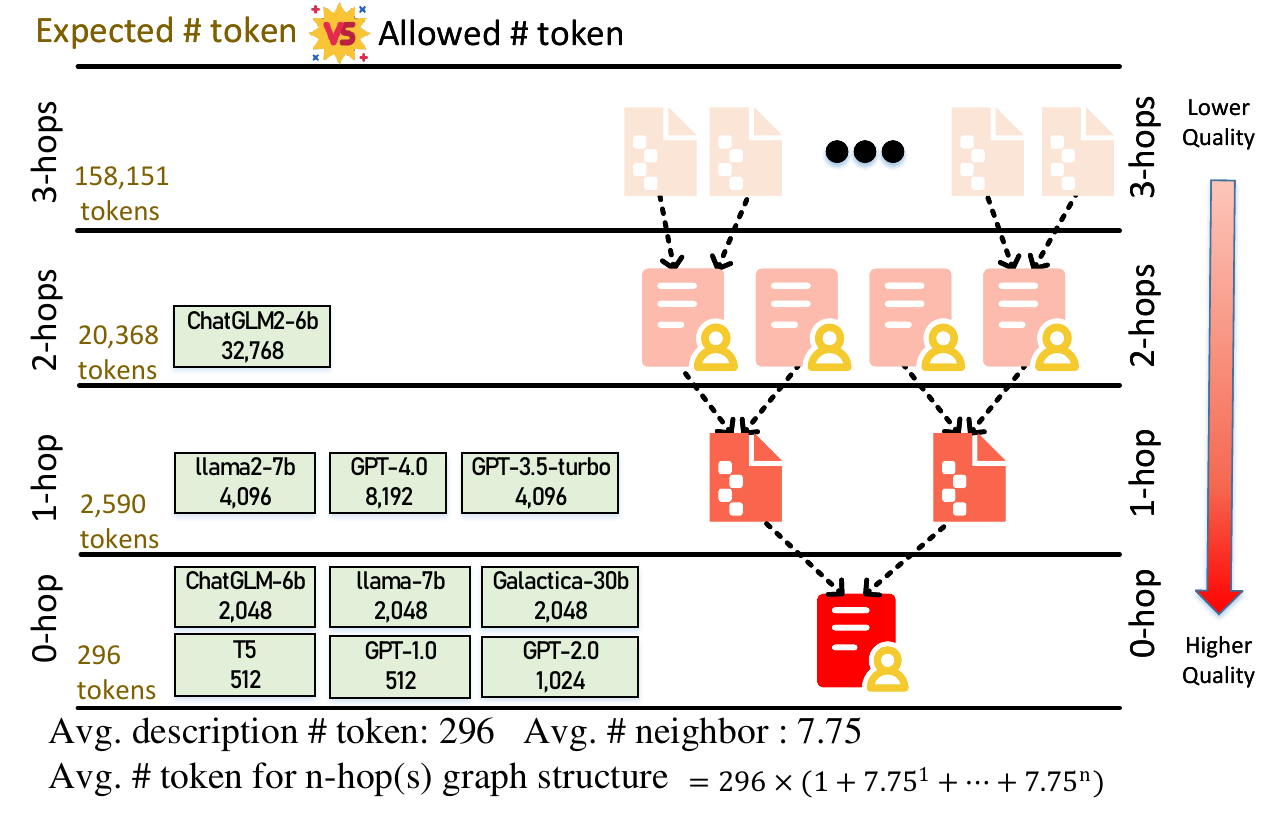}
\caption{Comparison between the token capacity of LLMs and the expected size (left part); Elicit the reasoning capacity of LLMs on the graph in a least-to-most manner (right part). } \label{fig2}
\end{figure}

Although exploring high-order descriptive information with LLMs is promising, integrating LLMs for structured graphs remains a challenge. First, text-based LLMs are ill-suited for processing graph-based information. Most existing methods convert graphical data into textual form using templates with sampling strategies~\cite{wang2023enhancing,andrus2022enhanced,wu2023exploring}, then feed texts into LLMs for downstream tasks. However, these methods impede the LLMs' global perspective on graphs, making it hard to fully utilize their reasoning skill in graph-based knowledge. 
Second, the limited context length of LLMs poses obstacles in effectively capturing graph-oriented information. The left side of Figure \ref{fig2} compares the maximum token capacity of LLMs and the expected length for describing a node in graphs based on different hops. It indicates that most LLM backbones even struggle to utilize merely 2-hop structural information from graphs.

To address these challenges, we propose \underline{G}raph-\underline{a}ware \underline{C}onvolut-ional \underline{LLM} (GaCLLM) to bridge the gap between text-based LLMs and graph-based information for recommendation. To adapt text-based LLMs with structured graphs, we develop a convolutional inference strategy inspired by Graph Convolution Networks (GCNs)~\cite{kipf2016semi}. Specifically, we deploy the LLM as an "aggregator" to assimilate neighbors within the graph iteratively, thus ensuring layer-by-layer information propagation throughout the graph structure. More importantly, the convolutional inference strategy can elicit the reasoning capacity of LLMs on the graph in a least-to-most manner~\cite{zhou2022least}, incrementally improving descriptions of users and items step by step, as shown in the right side of Figure \ref{fig2}. 
To alleviate the problem of excessive context length for LLMs, we segment the overload of multi-hop graph descriptions into multiple steps of queries. Particularly, we integrate the descriptions of one-hop neighbors at each time, therefore significantly reducing the required context length for LLMs. The key contributions include the following points:
\begin{itemize}
\item We propose a graph-aware convolutional inference strategy that uses LLMs to enhance each description progressively, to alleviate the challenges of the ill-fitness between LLMs and graphs, as well as the constrained context length in LLMs.
\item We underscore the significance of integrating graph-based knowledge into LLMs through data analysis. Compared to GCN-based methods, our proposed GaCLLM shows its advantage in leveraging the extensive knowledge base when reasoning on graphs. 
\item We evaluate our model on three real-world datasets. The experimental results show that our model consistently outperforms state-of-the-art methods. Comprehensive ablation experiments prove the effectiveness of the proposed method and justify the foundational motivations of our work.
\end{itemize}

\section{RELATED WORK}\label{Sec_RELATEDWORK}

\subsection{Graph-based Recommendation} In recent years, graph-based techniques have gained popularity in recommender systems~\cite{kipf2016semi}. They utilize deep neural networks to capture the complex interactive patterns between users and items of graph-structured data~\cite{jin2020multi,ying2018graph}. For example, Wang et al.~\cite{wang2019knowledge} adopt GCN to discover the high-order structure information in the knowledge graph to enrich users' preference modeling. LightGCN~\cite{he2020lightgcn} investigates the key components in GCNs and makes them more concise and appropriate for collaborative filtering, thus it gains popularity on recommendation tasks due to its simplicity and effectiveness. As a result, many studies adopt LightGCN using techniques such as contrastive learning~\cite{yu2022graph}, neighborhood-structure~\cite{lin2022improving}, and self-supervised learning~\cite{wu2021self}. However, these methods primarily focus on aggregating the embeddings of different nodes in the graph. They struggle to extract insights from text-level descriptions for semantic reasoning and description improvement.

\subsection{Large Language Models for Recommendation}

With the popularity of LLMs~\cite{touvron2023llama, brown2020language} in the field of natural language processing, there is growing interest in leveraging LLMs to enhance the efficacy of recommender systems~\cite{wu2023survey}. According to the roles of LLMs in recommendation tasks, existing methods can be primarily categorized into two groups:  LLM-as-predictor methods and LLM-as-extractor methods.

LLM-as-predictor methods, also known as generative recommendation, adopt the LLM as a predictor to directly generate recommendation results for users. These methods usually convert recommendation tasks into natural language processing tasks. Subsequently, techniques such as in-context learning~\cite{hou2023large,dai2022can}, prompt tuning~\cite{kang2023llms}, and instruction tuning~\cite{zhang2023recommendation} are employed to trigger LLMs for the direct generation of recommendation results.
 For example, \citet{zhang2023recommendation} employ instruction tuning in the prompt to inform LLMs about the requirements of the
recommendation task to better trigger LLMs to obtain satisfactory results. In contrast, LLM-as-extractor methods use the LLM as a feature extractor for downstream recommendation tasks. These methods boost the quality of recommendations due to the external knowledge and reasoning capacity of LLMs~\cite{liu2023pre}. Specifically, these methods aim to capture contextual information with superior efficacy, enabling a precise understanding of user profiles~\cite{zheng2023generative,du2023enhancing}, item descriptions~\cite{liu2023first}, and other textual data~\cite{geng2022recommendation}. For example, \citet{zheng2023generative} propose to fine-tune the LLM backbone based on the pairs of resumes and job descriptions if their candidates and employers reach an agreement, and then improve the user resume for the downstream recommendation task.
 However, current methods that prompt LLMs with raw texts ignore structured knowledge and easily lead to hallucinations, e.g., inconsistency between generation and users’ behaviors. To this end, we incorporate graph information into LLMs, to predict the missing descriptions of users and items for accurate recommendation results.


\subsection{Large Language Models in Graphs } 
Due to the availability and effectiveness of graph information, combining LLMs with graph data has emerged as a promising direction. According to the manner of how to explore graphical information with LLMs, existing methods can be primarily categorized into two groups: supervised methods and unsupervised methods.

Supervised methods typically adapt LLMs for graph-aware tasks with encoding strategies~\cite{chen2023exploring,zhang2021greaselm} and training strategies~\cite{sun2021ernie,yasunaga2022deep}. For example, \citet{chen2023exploring} use an LLM to encode the text information, and then integrate these encoding vectors into initial node features for graph models. \citet{sun2021ernie} propose a training objective by integrating entities and relations in the graph directly into the training data. However, these methods rely on compressing graph-based knowledge into the LLM’s parameters via supervision, leaving the reasoning capability of LLMs unexplored for graphs. Unsupervised methods~\cite{wang2023enhancing,andrus2022enhanced,wu2023exploring} attempt to transfer graph information into textual information with templates and sampling strategies for LLMs. For example, \citet{andrus2022enhanced} transfer knowledge graphs into text sentences based on prompts, then feed them into LLM to enhance story comprehension. 
\citet{wu2023exploring} sample semantic information in heterogeneous information networks based on meta-path, and leverage LLM-based recommender to understand such semantic information for job recommendation. 
However, the limitation of these methods lies in their reliance on templates and sampling strategies to generate text information in one shot, which lacks the global view of the graph and fails to elicit LLMs' reasoning capacity on graph-based knowledge inference. To this end, we propose to utilize LLMs to explore graphs in a least-to-most manner, therefore propagating information progressively in the graph with a limited token requirement.

\begin{figure*} \centering
 \includegraphics[width=1.0\textwidth]{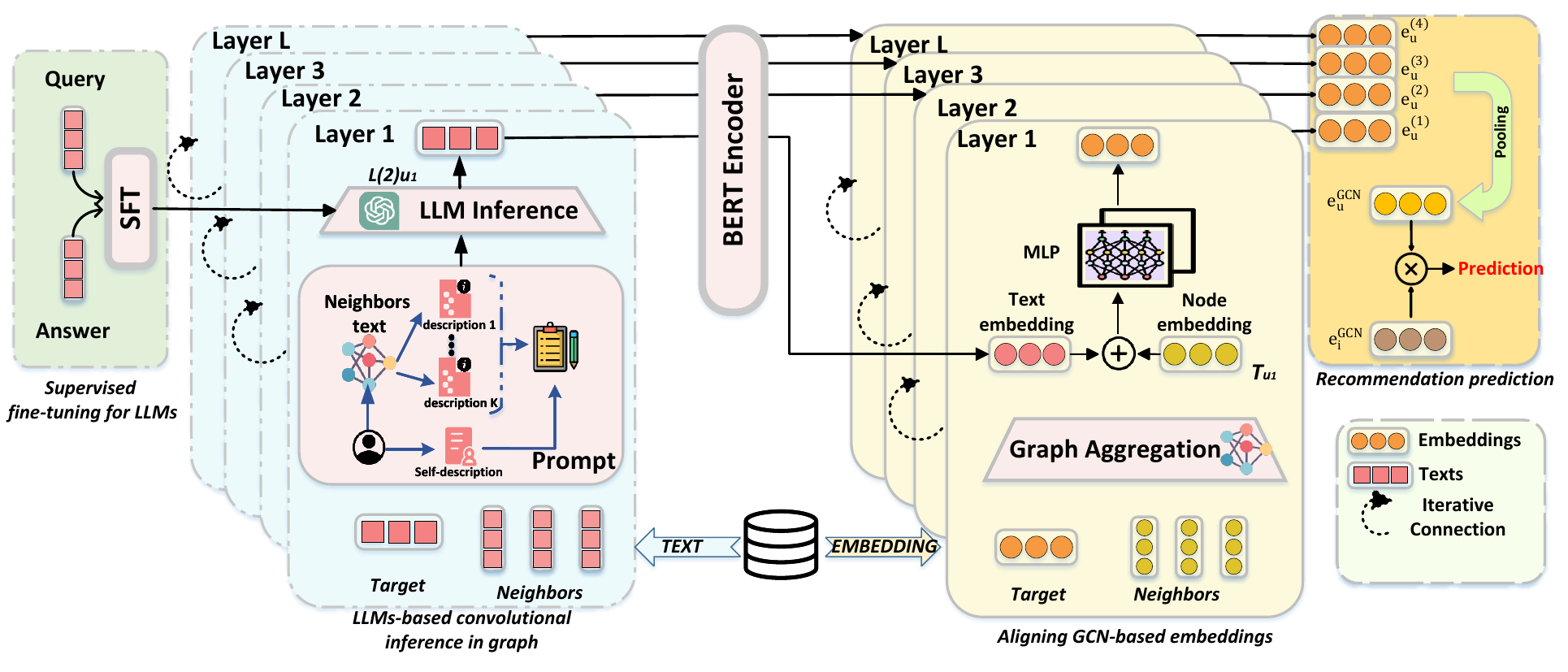}
\caption{The architecture of the proposed method GaCLLM. } \label{fig3}
\end{figure*}

\section{Preliminaries}
In this section, we first introduce the problem definition in this paper. Then, we introduce the basics of GCN-based methods for recommender systems.

\subsection{Problem Definition}
We denote $\mathcal{U}=\{u_1,\cdots,u_N\}$ and $\mathcal{I}=\{i_1,\cdots,i_M\}$ as the sets of users and items, where $N$ and $M$ denote the number of users and items respectively. The interaction records (e.g., purchase, click) between users and items can be denoted as an interaction matrix $\mathcal{R}\in\mathbb{R}^{N\times M}$ defined as follows:
\begin{equation*}
\mathcal{R}_{ui}=\begin{cases} 1  &\text{if user $u$ interacted with the item $i$},\\
0  &\text{otherwise}.  \\
\end{cases}
\end{equation*}
We also possess the description information of both users and items. 
We denote the description of user $u$ as $\mathcal{T}u = [w_1, \cdots, w{l_u}]$ and the description of item $i$ as $\mathcal{T}i = [w_1, \cdots, w{l_i}]$, with their lengths being $l_u$ and $l_i$ respectively. Here, $w_k$ represents the $k$-th word of a description. For example, in online recruitment scenarios, we have resumes as user descriptions and job descriptions as item descriptions, which are compulsory submissions in the platform. 

In this paper, our goal is to recommend appropriate items to users. To achieve this, we aim to learn a matching function $g(u, i)$ using the interaction records $\mathcal{R}$ and the provided textual descriptions of users and items. Our task is to recommend $K$ items that a user is most likely to prefer (aka top-$n$ recommendation).

\subsection{GCN Brief}

LightGCN~\cite{he2020lightgcn} has gained popularity in recommendation tasks due to its efficacy and efficiency. With the delicated design, LightGCN presents a simplified version of GCNs with the following aggregation layers:

\begin{equation*}
\bm{e}_{u}^{(l+1)} =AGG(\{ \bm{e}_{i}^{(l)}:i\in \mathcal{N}_{u} \}) = \sum_{i\in \mathcal{N}_{u}}\frac{1}{\sqrt{|\mathcal{N}_{u}||\mathcal{N}_{i}|}}\bm{e}_{i}^{(l)},
\end{equation*}
\begin{equation*}
\bm{e}_{i}^{(l+1)} =AGG(\{ \bm{e}_{u}^{(l)}:u\in \mathcal{N}_{i} \}) = \sum_{u\in \mathcal{N}_{i}}\frac{1}{\sqrt{|\mathcal{N}_{u}||\mathcal{N}_{i}|}}\bm{e}_{u}^{(l)}.
\end{equation*}
The representations in the first layer are initialized with the embeddings derived from the IDs of users and items, denoted as $\bm{e}^{(1)}_u\in \mathbb{R}^d$ and $\bm{e}^{(1)}_i\in \mathbb{R}^d$ for user $u$ and ID embedding of item $i$ respectively, with $d$ as the dimension of latent embedding space. The final GCN-based embeddings of users and items are obtained by fusing their multi-layer embeddings, e.g., $\bm{e}_{u}=Average(\bm{e}_{u}^{(1)},\cdots,\bm{e}_{u}^{(L)})$ and $\bm{e}_{i}=Average(\bm{e}_{i}^{(1)},\cdots,\bm{e}_{i}^{(L)})$, where $L$ denotes the total number of layers in the model. 
Several graph-based methods extend the LightGCN with contrastive learning~\cite{yu2022graph}, self-supervised learning~\cite{wu2021self}, etc. In summary, the key idea of GCNs-based methods is aggregating the neighborhood embeddings for each node layer by layer, thus the structural information can propagate over the graph.

\section{The proposed method} \label{Sec_Model}

The overall architecture of our proposed GacLLM is illustrated in Figure \ref{fig3}. It has four main modules. First, we conduct the supervised fine-tuning (SFT) for the LLM to activate its power in the task-related domain. Second, we propose an LLM-based graph-aware convolutional inference strategy to enhance the description of users and items progressively. Third, we align and fuse the generated description into graph-based embeddings of users and items. Last, we introduce the objective function and learning process of our proposed method.

\subsection{Supervised Fine-tuning for LLMs}
To harness the full potential of the LLM in understanding the task-related domain, we fine-tune the LLM with domain-specific data in the first stage. This involves the training of the LLM with descriptions from matched user-item pairs, in which case LLMs learn to align descriptions of users and items. Specifically, we employ the prompt template: "\textbf{Query}: Given an item's description, generate a user's description that fits it. The item's description is [Item Description]. \textbf{Answer}:", where [Item Description] is replaced with the actual description for the item. Symmetrically, we construct the prompt template for items. The optimization process  involves minimizing the negative log-likelihood loss for these templates as follows:
\begin{align*}
\mathcal{L}_{sft} =-\log \text{Pr}(T_{\text{Answer}}|T_{\text{Query}})  =-\sum_{k=1}^{|T_{\text{Answer}}|}\log\text{Pr}(v_k|v_{<k},T_{\text{Query}}),
\end{align*}
where $v_k$ denotes the $k$-th word in Answer sentence $T_{\text{Answer}}$, and $\text{Pr}(T_{\text{Answer}}|T_{\text{Query}}) $ denotes the generation probability for the produced answer with a given query. For the sake of time and computational efficiency, we adopt the LoRA strategy~\cite{hu2022lora} for parameter-efficient learning.

\subsection{LLM-based Convolutional Inference in Graph}
\subsubsection{Graph construction.} To explore the structured graph with high-order descriptive texts for LLMs, we propose to organize the descriptions of users and items into a unified graph $\mathcal{G}=(\mathcal{V},\mathcal{E})$. Specifically, the nodes $\mathcal{V}$ in the graph represent users and items,  i.e., $\mathcal{V}=\{{u}|u\in \mathcal{U}\}\cup\{{i}|i\in\mathcal{I}\}$. The edges $\mathcal{E}$ are constructed by the interactions between users and items $\mathcal{R}\in\mathbb{R}^{N\times M}$, i.e., $\mathcal{E}=\{(u,i)|\mathcal{R}_{u,i}=1\}$. Each node in the graph has a textual description, such as a summary of a movie, a profile of a user on social networks, and a resume of a job seeker.

\subsubsection{LLM-based convolutional inference strategy.} Recognizing the extensive knowledge, powerful text comprehension, and reasoning capabilities of LLMs, we propose an LLM-based convolutional inference strategy to explore high-order relations of descriptions in the graph. To improve the quality of descriptions for users, we leverage LLMs to rewrite the user's raw description $\mathcal{T}_{u}$ considering the descriptions of items that the user has engaged given by,
\begin{align*}
\mathcal{T}_{u}' = \mathtt{LLM}(\mathtt{PROMPT}_\mathtt{user}( \mathcal{T}_{u},\{ \mathcal{T}_{i}:(u,i)\in \mathcal{E} \})),
\end{align*}
where $\mathtt{PROMPT}_\mathtt{user}$ denotes the template for users' description generation (rewriting).
To improve the quality of descriptions for items, symmetrically, we leverage LLMs to rewrite descriptions of the item's raw description $\mathcal{T}_{i}$ considering the descriptions of users who engaged this item given by,
\begin{align*}
\mathcal{T}_{i}' =\mathtt{LLM}(\mathtt{PROMPT}_\mathtt{item} ( \mathcal{T}_{i},\{ \mathcal{T}_{u}:(u,i)\in \mathcal{E} \})),
\end{align*}
where $\mathtt{PROMPT}_\mathtt{item}$ denotes the template for items' description generation (rewriting). In real-world scenarios, the design of a prompt template varies with the tasks. In this paper, we focus on job and social recommendation tasks, the details of prompt templates for these two tasks are shown in Prompt1 and Prompt2.

To make text-based LLMs effectively explore the structured graph, we iteratively leverage LLMs to enhance the descriptions of nodes (users and items) step by step. Specifically, we set the first layer descriptions of users $\{\mathcal{L}_{u}^{(1)}|u \in \mathcal{U}\}$ by raw texts submitted by users, i.e., $\mathcal{L}_{u}^{(1)}=\mathcal{T}_{u}$. And we set the first layer descriptions of items $\{\mathcal{L}_{i}^{(1)}|i \in \mathcal{I}\}$ by raw texts submitted by item providers, i.e., $\mathcal{L}_{i}^{(1)}=\mathcal{T}_{i}$.  We position the LLM as an "aggregator" in the graph convolutional processing, boosting its analytical capability for graph-based knowledge inference through a stepwise progression.
\begin{equation}\label{eq10}
\mathcal{L}_{u}^{(l+1)} = \mathtt{LLM}(\mathtt{PROMPT}_\mathtt{user}( \mathcal{L}_{u}^{(l)},\{ \mathcal{L}_{i}^{(l)}:(u,i)\in \mathcal{E} \})),
\end{equation}
\begin{equation}\label{eq11}
\mathcal{L}_{i}^{(l+1)} =\mathtt{LLM}(\mathtt{PROMPT}_\mathtt{item} ( \mathcal{L}_{i}^{(l)},\{ \mathcal{L}_{u}^{(l)}:(u,i)\in \mathcal{E} \})),
\end{equation}
where $\mathcal{L}_{u}^{(l+1)}$ and $\mathcal{L}_{i}^{(l+1)}$ denote the description of users and items at $(l+1)$-th layer after $l$ iterations of rewriting, which can capture $l$-hop descriptive information in the graph. After $L$ iterations of the LLM-based convolutional inference strategy, we obtain different layers of user and item descriptions for both users and items. 

\definecolor{myboxcolor2}{RGB}{191,128,191}
\newtcolorbox{mybox3}[1]{
  colback=myboxcolor2!5!white,
  colframe=myboxcolor2!75!black,
  fonttitle=\bfseries,
  title=#1,
  halign title=center, 
  halign=flush left,    
  left=1mm,
  right=1mm
}

\begin{mybox3}{Prompt1: Job Recommendation}
\textit{\textbf{For user resumes}:\newline
Please make appropriate improvements and revisions
to the user’s resume by inferring from his/her resume
and his interested job descriptions to generate a more concise
resume. The user’s resume is: [{\color{blue}Resume content}]. The job
descriptions that interest the user are: [{\color{blue}Job Description
1, Job Description 2, . . . , Job Description K}].
\newline
\textbf{For job descriptions}:\newline
Please make appropriate improvements and revisions
to the job description by inferring from the original description
and resumes of users who are interested in the job, generating a more concise
job description. The original description is: [{\color{blue}Job Description}]. The resumes of users who
are interested in the job are: [{\color{blue}Resume 1, Resume 2, . . . , Resume K].}}
\end{mybox3}

\begin{mybox3}{Prompt2: Social Recommendation}
\textit{Please make appropriate improvements and revisions of the user introduction based on commonalities of his/her self-description and friends' description. The user self-description is: [{\color{blue}User description}]. His/her friends' descriptions are: [{\color{blue}Friend description 1, …, Friend description K}].}
\end{mybox3}




\subsubsection{Strengths of the LLM-based convolutional strategy.} 
We claim the proposed LLM-based convolutional strategy has two main strengths for recommendation.

\smallskip\noindent\textbf{Least-to-Most Reasoning}: LLMs employ a sequential approach in their iterative aggregation process, gradually building higher-quality descriptions of users and items in a least-to-most manner. The proposed strategy segments the graph into a hierarchical structure as shown in Figure \ref{fig2}, highlighting LLMs' adaptability to complex graph-based knowledge.

\begin{figure} \centering
 \includegraphics[width=8.5cm]{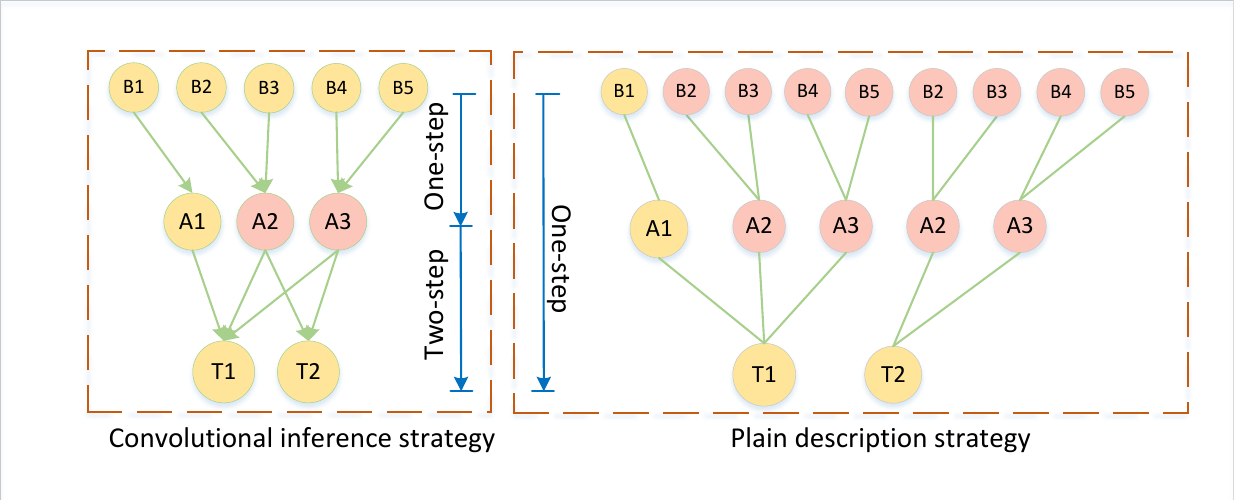}
\caption{Different ways for token usage in LLMs.}\label{fig_Efficiency}
\end{figure}

\smallskip\noindent\textbf{Effectiveness and Efficiency in Token Usage}: Compared to describing all node descriptions related to the target node in a plain way, the proposed convolutional inference strategy can achieve both effectiveness and efficiency in token usage in two manners. First, it can effectively capture graph-related information with a limited context length of LLMs. Specifically, the proposed strategy breaks down the task of description enhancement into multiple steps. Each step (layer) only integrates the descriptions of one-hop neighbors for the target node, thus leading to a significant decrease in the number of tokens needed. Second, the proposed strategy can efficiently alleviate the redundancy of describing the graph for target nodes. 
Specifically, we compare the number of nodes required to describe the $L$-hop graph-based information for each node in the graph.
Our proposed strategy needs to incorporate  $O(|\mathcal{G}|\cdot|\mathcal{N}|\cdot L)$ nodes into LLMs, while the plain description strategy needs to incorporate $O(|\mathcal{G}|\cdot(1 + \cdots+|\mathcal{N}|^{ L}))$ nodes into LLMs, where $|\mathcal{G}|$ denotes the number of nodes in the graph and $|\mathcal{N}|$ denotes the average number of neighbors for each node.
Therefore, our proposed strategy can decrease the description of overlapped nodes (pink ones) compared to plainly describing all node descriptions as shown in Figure \ref{fig_Efficiency}.  As a result, the effectiveness and efficiency of the convolutional inference strategy contribute to the real-world applicability of the proposed method.



%

\subsection{Aligning GCN-based Embeddings for Recommendation}
\subsubsection{Aligning GCN-based embeddings of users and items.}
To bridge the text information and structural information in the graph for recommendation, we propose to align the descriptions of users and items with their embeddings in a unified manner. Specifically, the $l$-th layer of GCN-based embeddings of users and items $\bm{e}_{u}^{(l+1)}$ and $\bm{e}_{i}^{(l+1)} $ can be formulated as follows:

\begin{align*}
\bm{e}_{u}^{(l+1)} = W_l\cdot \text{contact}[\sum_{(u,i)\in \mathcal{E}}\frac{1}{\sqrt{|\mathcal{N}_{u}||\mathcal{N}_{i}|}}\bm{e}_{i}^{(l)};encoder(\mathcal{L}_{u}^{(l)})], \\
\bm{e}_{i}^{(l+1)} = W_l\cdot \text{contact}[\sum_{(u,i)\in \mathcal{E}}\frac{1}{\sqrt{|\mathcal{N}_{u}||\mathcal{N}_{i}|}}\bm{e}_{u}^{(l)};encoder(\mathcal{L}_{i}^{(l)})].
\end{align*}
Here, the first layer of GCN-based embeddings represents the initial step for both users' and items' representations. In this layer, each user and item is associated with the embedding from a specific ID, i.e.,  $e^{(0)}_u\in \mathbb{R}^d$ and $e^{(0)}_i\in \mathbb{R}^d$. $W_l\in\mathbb{R}^{2d \times d}$ denotes the transformation mapping matrix for the $l$-th layer. To model the enhanced description text that is associated with a user or an item, we propose to encode them into the constant text-based embedding by $encoder(\cdot)$. Specifically, we first keep the original text order and pad a unique token $[CLS]$ before the text, then we feed the combined sequence into the simbert-base-chinese\footnote{https://huggingface.co/WangZeJun/simbert-base-chinese} model and use the output of token $[CLS]$ as the semantic embeddings of the descriptive text.

To make use of the descriptions of users and items across different layers,  we further combine their embeddings from each layer to produce the final embeddings of users and items, i.e.,
\begin{equation*}
\bm{\widetilde{e} }_{u} =\frac{1}{L}\sum\nolimits_{l=1}^{L} \bm{e}_{u}^{(l)} ;\quad
\bm{\widetilde{e}}_{i} =\frac{1}{L}\sum\nolimits_{l=1}^{L} \bm{e}_{i}^{(l)}.
\end{equation*}

\subsubsection{Objective function.}
To measure the matching scores between users and items for final predictions, we propose to compute the inner product of their representations, i.e.,
\begin{equation*}\label{loss_rec}
\hat{R}_{u,i} = <\bm{\widetilde{e}}_{u},\bm{\widetilde{e}}_{i}>,
\end{equation*}
where $<\cdot,\cdot>$ denotes the operation of inner product. It produces a score or probability of item $i$ that user $u$ will engage. For the model training process, we use the pairwise loss to define the recommendation objective function as follows:
\begin{equation*}
     \max\limits_{\Theta} \sum_{(u,i,j)\in \mathcal{D}}\log
\sigma(\hat{R}_{u,i} - \hat{R}_{u,j})  - \lambda ||\Theta||^2,
\end{equation*}
where the train set $\mathcal{D} = \{(u,i,j)\}$ consists of triplets with a user $u$, an item $i$ with positive feedback from user $u$, and an item $j$ with negative feedback from user $u$.  $\Theta$ denotes all parameters that need to be trained in the proposed model, and $\lambda$ is the regularization coefficient of L2 norm $||\cdot||^2$.

\section{Experiment} \label{Sec_Experiments}
In this section, we aim to evaluate the performance and effectiveness of the proposed method GaCLLM. Specifically, we conduct a group of experiments to study the following research questions:
\begin{itemize}
\item \textbf{RQ1}: Whether the proposed method outperforms state-of-the-art methods for recommendation? 
\item \textbf{RQ2}: Whether the proposed method benefits from the LLM-based convolutional inference and text-graph alignment modules? 
\item \textbf{RQ3}: How and to what extent can the proposed method enhance the descriptions of users and items and contribute to the improvement of recommendation results?
\item \textbf{RQ4}: How do different configurations of key hyper-parameters and module implementation impact the performance of the proposed methods?

\end{itemize}
\begin{table}[]
\caption{Statistics of the experimental datasets} \centering
\label{Table_es}
\begin{tabular}{cccc}
\toprule
 \textbf{Dataset} & \# Users & \# Items & \# Interaction  \\
\midrule
 Designs &  12,290 & 9,143  & 166,270 \\\midrule
 Sales   &  15,854 & 12,772 & 145,066 \\ \toprule
 Pokec    &  6,240 (Male) & 6,213  (Female) & 104,152   \\
\bottomrule
\end{tabular}
\end{table}
\begin{table*}[htbp!]
  \centering
  \fontsize{10}{12}\selectfont
  \caption{Performance of the proposed GaCLLM and baseline methods for recommendation.  $*$ indicates that the improvements are significant at the level of 0.05  with a paired t-test.}
  \label{tab:RSComparison}
\begin{tabular}{c|cc|cc|cc|cc}
\toprule
 \multirow{2}[3]{*}{\textbf{Models}} & \multicolumn{2}{c|}{\textbf{Designs}} & \multicolumn{2}{c|}{\textbf{Sales}} & \multicolumn{2}{c|}{\textbf{Pokec-M}} & \multicolumn{2}{c}{\textbf{Pokec-F}} \\ \cmidrule{2-9}
 & \multicolumn{1}{c}{MAP@5} & \multicolumn{1}{c}{NDCG@5}& \multicolumn{1}{c}{MAP@5} & \multicolumn{1}{c}{NDCG@5} & \multicolumn{1}{c}{MAP@5} & \multicolumn{1}{c}{NDCG@5}  & \multicolumn{1}{c}{MAP@5} & \multicolumn{1}{c}{NDCG@5}\\ \midrule
SGPT-ST       & 0.0651          & 0.1042          & 0.0491          & 0.0861          & 0.0710& 0.0980
          & 0.0724& 0.1013      \\
MF            & 0.2081          & 0.3182          & 0.0957          & 0.1751          & 0.2616          & 0.3876          & 0.2639          & 0.3838          \\
NCF           & 0.2100          & 0.3258          & 0.1468          & 0.2678          & 0.2930          & 0.4273          & 0.2969          & 0.4270          \\\midrule
LightGCN-text    & {\ul 0.2940}    & {\ul 0.4697}    & 0.1658          & 0.3001          &{\ul 0.3294}           &  {\ul0.4676 }         & {\ul 0.3293}    & 0.4664          \\
SimGCL-text   & 0.1471          & 0.2277          & 0.0921          & 0.1658          & 0.3093         & 0.4459          & 0.2940          & 0.4235          \\
UltraGCN-text & 0.2639          & 0.4258          & 0.1469          & 0.2725          & 0.3204    & 0.4623    & 0.3263          & {\ul 0.4691}    \\
SGL-text      & 0.2769          & 0.4418          & 0.1431          & 0.2567          & 0.3012          & 0.4394          & 0.3047          & 0.4385          \\\midrule
LLM-CS       & 0.2669          & 0.2190          & 0.1530          & 0.2803          & 0.2527          & 0.3468          & 0.2569          & 0.3478          \\
LLM-TES      & 0.2208          & 0.3478          & 0.1520          & 0.2797          & 0.2571          & 0.3512          & 0.2593          & 0.3517          \\
LGIR          & 0.2898          & 0.4616          & {\ul 0.1694}    & {\ul 0.3103}    & 0.3081          & 0.4183          & 0.3245	 &0.4390\\\midrule
\textbf{GaCLLM} & \textbf{0.3060*} & \textbf{0.4925*} & \textbf{0.1750*} & \textbf{0.3234*} & \textbf{0.3446*} & \textbf{0.4797*} & \textbf{0.3461*} & \textbf{0.4798*}\\
Improve. &4.06\%&	4.85\%&	3.32\%&	4.21\%&	4.60\%&	2.60\%&	5.10\%&	2.28\%\\\bottomrule

\end{tabular}
\end{table*}

\subsection{Experimental Setup}
\subsubsection{Datasets.}
In this paper, we investigate two different scenarios for recommendation: job recommendation and social recommendation.
For job recommendation, we evaluate the proposed model using two real-world datasets sourced from an online recruiting platform within the designer and sales industries. These two datasets encompassed extensive user-job interactions. Furthermore, they include textual document information, comprising user resumes and job descriptions.  For social recommendation, we evaluate the proposed method using one real-world dataset sourced from an online social platform, which is the most popular online social network for inviting someone on a date in Slovakia. This dataset contains the friend relation between users and the self-descriptions of users, we focus on recommending friends of the opposite sex to users (i.e., Pokec-M and Pokec-F for male and female users, respectively). The characteristics of these datasets are summarized in Table \ref{Table_es}.

\subsubsection{Evaluation Methodology and Metrics.}

We randomly split the interaction/relations records into three equal sets, namely the training set, the validation set, and the test set. For performance evaluation, we utilize two well-recognized top-$n$ recommendation metrics, namely, mean average precision ($MAP@n$) and normalized discounted cumulative gain ($NDCG@n$), where the value of $n$ was determined empirically and set to 5. Following previous studies \cite{du2023enhancing,yang2022modeling}, We select 20 negative instances for every positive instance for testing, and the experimental results represent the average of five runs with different random initializations of model parameters.

\subsubsection{Baselines.}
We took the following state-of-the-art methods as the baselines, including content-based methods~\cite{muennighoff2022sgpt,Jiang2020}, collaborative filtering methods~\cite{BPR,he2017neural}, graph-based methods~\cite{he2020lightgcn,yu2022graph,mao2021ultragcn,wu2021self}, and LLMs based method~\cite{du2023enhancing,chen2023exploring}. 

Specifically,
\textbf{SGPT-BE}~\cite{muennighoff2022sgpt} applies GPT models as {B}i-{E}ncoders for asymmetric search.
\textbf{PJFFF}~\cite{Jiang2020} fuses the representations for the explicit and implicit intentions of users and employers by the historical application records.
\textbf{MF}~\cite{koren2009matrix} learns low-dimensional representations of users and items by reconstructing their interaction matrix based on the point loss.
\textbf{NCF}~\cite{he2017ncf} enhances collaborative filtering with deep neural networks, which adopt an MLP  to explore the non-linear interaction between user and item.
 For a fair comparison, we enhance all graph-based methods (LightGCN, UltraGCN, SimGCL, and SGL) with text information, i.e., we use simbert-base-chinese to encode text and then concatenate it with ID embeddings with an MLP $f(\cdot): \mathbb{R}^{2d}\to \mathbb{R}^d$.
\textbf{LightGCN-text}~\cite{he2020lightgcn} is the text-based version of LightGCN, which simplifies the vanilla GCN's implementation to make it concise for collaborative filtering.
\textbf{UltraGCN-text}~\cite{mao2021ultragcn} skips infinite layers of message passing of GCN for efficient recommendation.
\textbf{SimGCL-text}~\cite{yu2022graph} adds uniform noises to the embedding of GCN and conducts contrastive learning for recommendation.
\textbf{SGL-text}~\cite{wu2021self} conducts the self-supervised learning on the user-item graph to improve the accuracy and robustness of GCNs for recommendation. 
\textbf{LGIR}~\cite{du2023enhancing}infers users’ implicit characteristics from their behaviors for resume completion and designs a GAN-based model for recommendation.
\textbf{LLM-CS}~\cite{chen2023exploring}  directly encodes text attributes into initial node features by LLMs for graph models.
\textbf{LLM-TES}~\cite{chen2023exploring} adopts LLMs to enhance text attributes and encode them into initial node features for graph models.

\subsubsection{Implementation Details.}
We adopt the ChatGLM2-6B\footnote{https://huggingface.co/THUDM/chatglm2-6b}~\cite{du2022glm} as the LLM backbone in all methods for a fair comparison. For LLM backbone fine-turning, we adopt the LoRA strategy~\cite{hu2022lora} with a learning rate of $1\times10^{-5}$, LoRA dimension of 128, a batch size of 2, 10000 train steps, and gradient accumulation of 1. To make a fair comparison between baseline methods and ours in the model training phase, all methods are optimized by the AdamW optimizer with the same latent space dimension (i.e., $768$), batch size (i.e., $1024$), and regularization coefficient (i.e., $1\times10^{-4}$). 

\subsection{Model Comparison (RQ1)}

\renewcommand{\arraystretch}{1.4}


Table \ref{tab:RSComparison} shows the performance of different methods for recommendation. To make the table more notable, we bold the best results and underline the best baseline results in each case.  From the experimental results, we can get the following conclusions:
\begin{itemize}
\item  First, the proposed method GaCLLM consistently outperforms all baseline methods in all cases, improving
the best baseline by an average of 4.46\%, 3.77\%, 3.60\%, and 3.69\% on these datasets. This demonstrates the effectiveness of the proposed method. 
\item Second, it is noteworthy that several methods that only model text (e.g., SGPT-BE) or interaction information (e.g., MF and NCF) demonstrate poor performance compared to the other hybrid approaches, which demonstrate the necessity of utilizing both text and interaction information. 
\item Third, the GCN-based methods (e.g., LGCN-text and UltraGCN-text) achieve better performance across baselines, indicating that extracting graph information as well as text information helps with accurate recommendation results. This verifies our motivation of exploring structural information in the graph with the help of LLMs to improve the quality of descriptions for recommendation. Some GCN-based methods (e.g., SimGCL-text) exhibit unsatisfactory performance among baselines, which is attributed to the unfitness of their frameworks w.r.t. text-aware information.
\item Last, it shows that adopting the LLM as the encoder (i.e., LLM-CS) and simple resume completion (i.e., LLM-TES) shows inferior performance among the baseline methods. The relatively better performance of LGIR indicates that inferring from users' behavior can alleviate the issues of fabricated and hallucinated generation of LLMs. 
\end{itemize}

\subsection{Ablation Study (RQ2)}

\renewcommand{\arraystretch}{1.4}
\begin{table}[tp]
  \centering
  \fontsize{9}{9}\selectfont
  \caption{ Performance of the variants for ablation studies}\setlength{\tabcolsep}{3pt}
\begin{tabular}{|c|cc|cc|cc|cc|}
\toprule
 \multirow{2}[3]{*}{\textbf{Models}} & \multicolumn{2}{c|}{\textbf{Designs}} & \multicolumn{2}{c|}{\textbf{Sales}}  \\ \cmidrule{2-5}
 & \multicolumn{1}{c}{MAP@5} & \multicolumn{1}{c}{NDCG@5}& \multicolumn{1}{c}{MAP@5} & \multicolumn{1}{c|}{NDCG@5} \\ \midrule
GaCLLM-RAW           & 0.2951          & 0.4717          & 0.1692          & 0.3082                 \\
GaCLLM-PLAIN & 0.2908          & 0.4655          & 0.1677          & 0.3080           \\
GaCLLM-noALIGN      & 0.2901          & 0.4654          & \textbf{0.1753}          & 0.3212             \\
GaCLLM               & \textbf{0.3060} & \textbf{0.4925} & {0.1750} & \textbf{0.3234} \\\bottomrule
\toprule
 \multirow{2}[3]{*}{\textbf{Models}} & \multicolumn{2}{c|}{\textbf{Pokec-M}} & \multicolumn{2}{c|}{\textbf{Pokec-F}} \\ \cmidrule{2-5}
 & \multicolumn{1}{c}{MAP@5} & \multicolumn{1}{c}{NDCG@5}& \multicolumn{1}{c}{MAP@5} & \multicolumn{1}{c|}{NDCG@5} \\ \midrule
GaCLLM-RAW                    & 0.3326          & 0.4655          & 0.3402          & 0.4678          \\
GaCLLM-PLAIN         & 0.3287          & 0.4612          & 0.3362          & 0.4672          \\
GaCLLM-noALIGN              & 0.3331          & 0.4687          & 0.3435          & 0.4780          \\
GaCLLM              & \textbf{0.3446} & \textbf{0.4797} & \textbf{0.3461} & \textbf{0.4798}\\\bottomrule
\end{tabular}\label{table_ablation} 
\end{table}

\begin{figure*} \centering
 \includegraphics[width=1.0\textwidth]{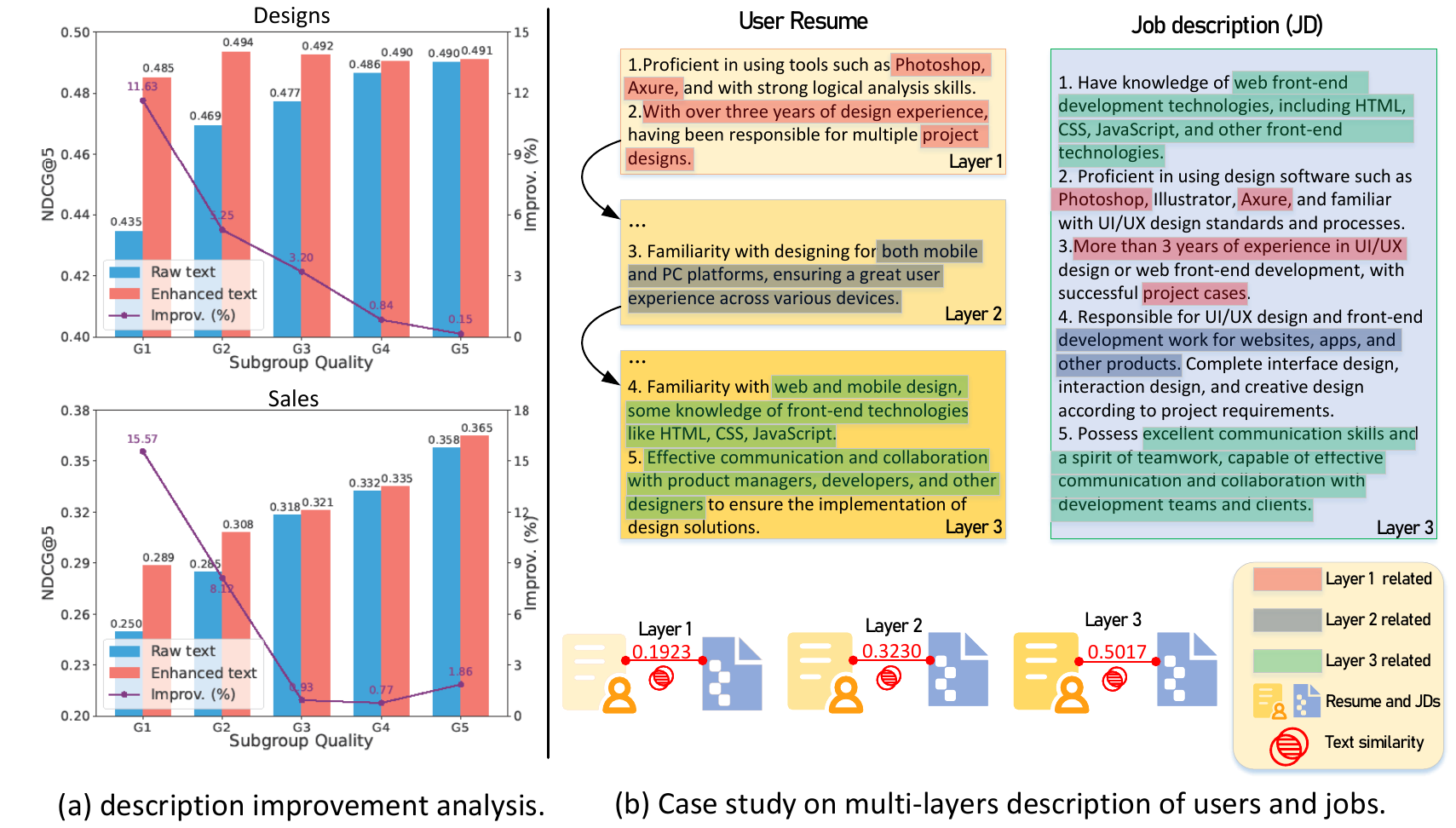}
\caption{(a) Performance across different quality levels 
of users’ raw descriptions for description improvement analysis. (b) for description improvement analysis.}\label{fig_mixture}
\end{figure*}

To verify the motivations presented in this paper, we consider different variants of the proposed method for comparison.
\begin{itemize}
\item[-]\textbf{GaCLLM-RAW}:
 It adopts the raw description into the proposed method GaCLLM for recommendation, i.e., $\mathcal{L}_{u}^{(L)}=\cdots=\mathcal{L}_{u}^{(1)} = \mathcal{T}_{u}$ and $\mathcal{L}_{i}^{(L)}=\cdots=\mathcal{L}_{i}^{(1)} = \mathcal{T}_{i}$.
\item[-]\textbf{GaCLLM-PLAIN}:
It adopts a template to describe all node descriptions related to the target node in a plain way, then feeds it into the LLM for description generation. For example, the template of the user resume enhancement can be written by "... The user's resume is: [Resume content]. The job descriptions that interest the user are: [Job Descriptions 1, which interests users with Resume $1$, ..., Resume $n_1$]; .... ;  [Job Descriptions $k$, which interests users with Resume $1$, ..., Resume $n_k$]". 
\item[-]\textbf{GaCLLM-noALIGN}:
 It discards the alignment of GCN-based embeddings for each layer, and it adopts the $L$-th layer descriptions to enhance the initial embeddings of GCN for recommendation.
\end{itemize}
Table \ref{table_ablation} shows the performance of ablation models, i.e. GaCLLM-RAW, GaCLLM-PLAIN, GaCLLM-noALIGN, and GaCLLM. From the experimental results, we can get the following conclusions:

\begin{itemize}
\item First, the proposed method GaCLLM consistently outperforms GaCLLM-RAW on all datasets, which indicates that the proposed LLM-based convolutional inference strategy can improve the quality of users' and items' descriptions and thus lead to more accurate recommendation results.
\item Second, GaCLLM significantly outperforms GaCLLM-PLAIN, and we believe this can be attributed to two key factors. On the one hand, GaCLLM-PLAIN can hardly understand the structured graph by simply describing it in a plain way; On the other hand, GaCLLM shows priority of eliciting the reasoning capacity of LLMs on graph-based knowledge inference in a least-to-most manner. 
\item  Third, the proposed method GaCLLM outperforms GaCLLM-noALIGN in most cases. It indicates that the aligning of both text-level and embedding-level representations can bridge their gap and make use of different layers of description generated by the LLM for recommendation.
\end{itemize}

\subsection{Description Improvement Analysis  (RQ3)}\label{sec_hyper}
The ablation study illuminates GaCLLM's strengths in enhancing user and item descriptions. It is interesting to investigate whether GaCLLM can capture our claims at the case level by employing the proposed LLM-based convolutional inference strategy.

\subsubsection{Subgroup analysis.} To investigate how and to what extent GaCLLM can enhance the descriptions of users and items, we conduct a comparison with the variant GaCLLM-RAW across different quality levels of users' raw descriptions. Specifically, users are equally divided into five groups $G1\sim G5$ based on their description length ascendingly (an illustrative example in Figure \ref{fig1} shows that the description length is highly correlated with its quality), and the recommendation performance of GaCLLM and GaCLLM-RAW was compared across these groups as shown in Figure \ref{fig_mixture} (a). First, GaCLLM consistently outperforms GaCLLM-RAW across all user subgroups, indicating that GaCLLM benefits from description enhancement across all quality levels. Second, GaCLLM exhibits more substantial improvements within the lower-quality user subgroups, underscoring the effectiveness of the proposed LLM-based convolutional inference strategy for description improvement as we claimed.
\subsubsection{Case study.} Fig.\ref{fig_mixture} (b) shows a case-level description improvement process in the Designs dataset. 
For page limitation, we only show the improvement process of the user's resume. We highlight content relevant to a target job among the user's resumes with different layers. First, with the growth of the layers, the user resume shows more related information to the target job, which indicates the necessity of eliciting the LLM's reasoning capacity on the graph in a least-to-most manner. Second, the text similarity between the user's resume and job description shows a remarkable improvement in the higher layer, based on which crafting users' and items' representations contributes to better recommendation results.

\renewcommand{\arraystretch}{1.4}

\begin{figure} \centering
 \includegraphics[width=8.5cm]{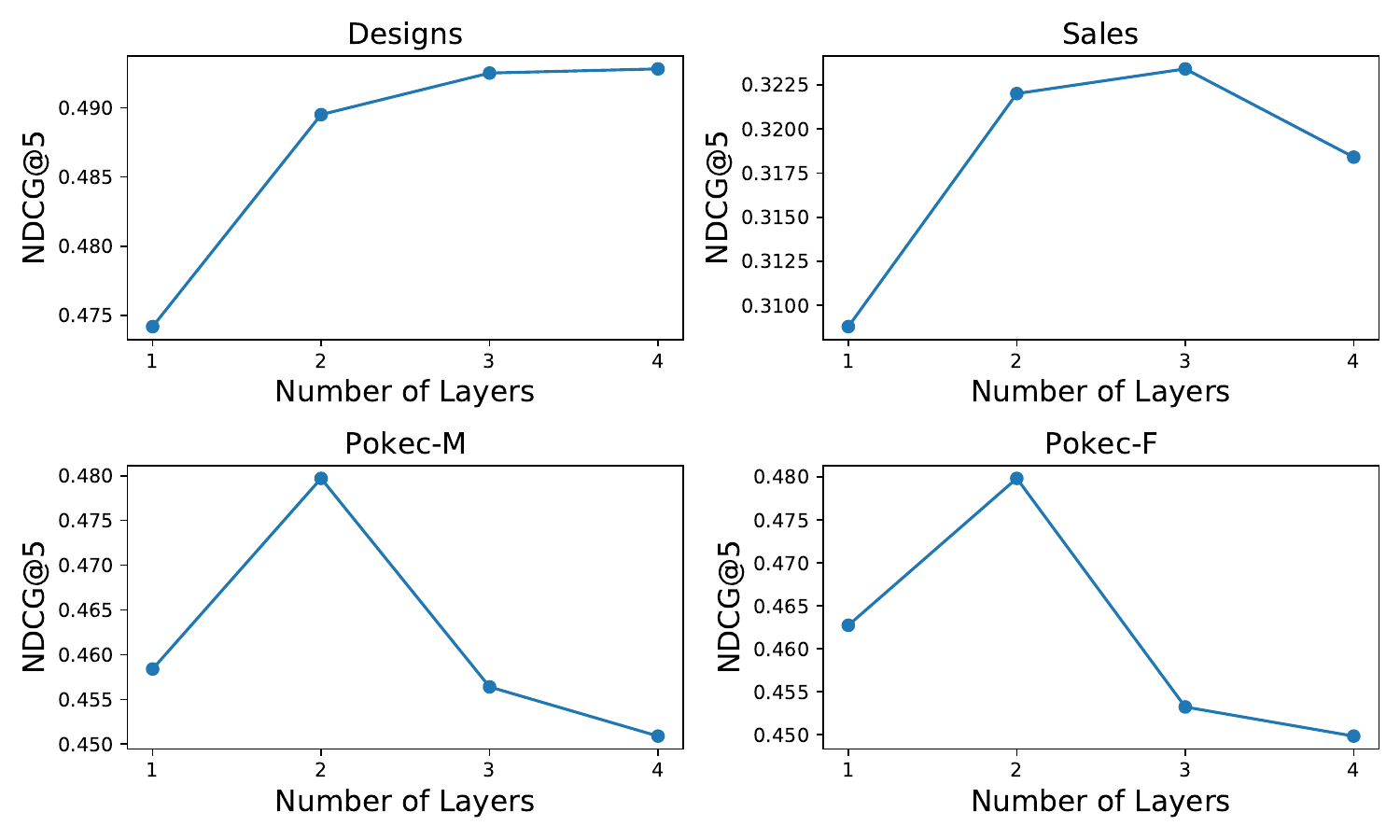}
\caption{Performance of the proposed method varying with different numbers of layers.}\label{fig_hyper}
\end{figure}

\begin{figure} \centering
 \includegraphics[width=8.5cm]{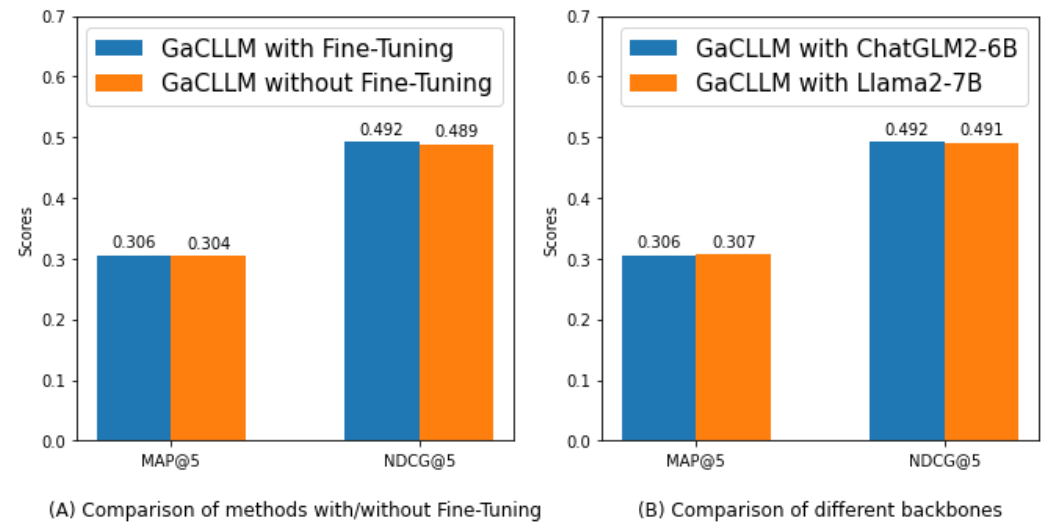}
\caption{Performance of the proposed method varying with LLMs backbone in Designs dataset.}\label{fig_backbone}
\end{figure}

\begin{table}[]
\caption{ Performance of the proposed method varying with text encoder in Designs dataset.} \centering
\label{Table_encoder} 
\fontsize{8}{8}\selectfont
\begin{tabular}{|c|c|c|}
\midrule
Encoder              & MAP@5  & NDCG@5    \\\midrule
simbert-base-chinese & 0.3060 & 0.4925  \\\midrule
ChatGLM2-transformer & 0.2722 & 0.4291 \\\midrule
\end{tabular}
\end{table}

\subsection{Hyper-Parameter Study (RQ4)}\label{sec_hyper}
{
We also evaluate the effects of different settings of the key hyper-parameters and modules on the performance of the proposed method,
including the number of layers, supervised fine-tuning module, LLM backbone, and text encoder.
\subsubsection{Number of layers}
We aim to explore how the number of layers affects the performance of our proposed method GaCLLM. As depicted in Figure \ref{fig_hyper}, we assess GaCLLM's performance under different layer configurations. We observe that the highest performance is achieved when layer numbers are $(4, 3, 2)$  for designs, sales, and pokec datasets, respectively. For real-world applications, we suggest adopting a grid search strategy as a practical approach to select the optimal layer numbers for implementing the GaCLLM method.
\subsubsection{Supervised fine-tuning study}
We explore whether the proposed method GaCLLM benefits from supervised fine-tuning of the LLM backbone. As illustrated in Figure \ref{fig_backbone} (A), we evaluate the GaCLLM variant both with and without supervised fine-tuning on the Designs dataset. Specifically, the experiment results show that fine-tuning LLMs shows limited improvement, which indicates that the performance improvement in the proposed method is NOT obtained directly from continued pre-training. In other words, the proposed method mostly benefits from the LLM-based convolutional inference strategy and GCN-based embedding alignment, which verified the motivation of the proposed method.  Although the proposed method GaCLLM achieves a limited improvement of supervised fine-tuning in these datasets, we believe that some scenarios may contain much extra domain-specific knowledge beyond LLMs’ public knowledge. Therefore,  we have made the fine-tuning process an optional setting, contributing to the adaptability of the proposed method.
\subsubsection{LLM backbone study}
We investigate how different backbones affect the performance of our proposed method GaCLLM. As shown in Figure \ref{fig_backbone} (B), we assess the performance of GaCLLM using the chatGLM2-6B and Llama2-7B backbones, both of which demonstrate comparable scales.  The experimental results indicate that GaCLLM achieved comparable performance with these two backbones, confirming the robustness of text generation through the convolutional inference strategy.
\subsubsection{Text-encoder study}
To bridge the gap between text information and embeddings, we employ Chinese-SIM-BERT to encode user and item text information into a latent space. It is interesting to investigate whether the LLM's transformer can be utilized as the text encoder for sentence representation encoding.
Table \ref{Table_encoder} reveals suboptimal performance with the ChatGLM2-transformer as a text encoder. This could be due to LLMs being primarily decoder-based models for autoregressive text generation. Therefore, we recommend employing a BERT-based model as the text encoder for practical applications.
}

\section{Conclusion} \label{Sec_Conclusion}
In this paper, we propose a graph-aware convolutional LLM method to improve the quality of textual information for recommendation. Our method successfully bridges the gap between text-based LLMs and graph-based multi-hop information. Specifically, we devise a convolutional inference strategy that iteratively employs LLMs to enhance the description of nodes in a least-to-most manner. Using this approach, we enable LLMs to infer description information from a graph-based perspective, allowing for propagation over the structure within a constrained token input size. Extensive experiments show that the proposed model consistently outperforms state-of-the-art methods. In addition, comprehensive ablation studies validate the effectiveness of our module design and underlying motivations. In the future, we will study how to use LLMs to explore the graph with heterogeneous relations, which can extract more fine-grained information for recommendation.
\bibliographystyle{ACM-Reference-Format}
\bibliography{sigconf}

\end{document}